\documentclass[runningheads]{llncs}
\usepackage[T1]{fontenc}
\usepackage{graphicx}
\usepackage{booktabs}
\usepackage[misc]{ifsym}
\newcommand{\corr}{(\Letter)}
\usepackage{tabularx}
\usepackage{multirow} % for \multirow command
\usepackage{subcaption}
\usepackage{multicol}
\usepackage{algorithm}
\usepackage{algpseudocode}
\usepackage{amsmath}
\usepackage{amsfonts}
\usepackage{amssymb}
\usepackage{comment}
\usepackage{hyperref}

\begin{document}

\title{Modeling Text-Label Alignment for Hierarchical Text Classification}

%\titlerunning{Underwater Basket Weaving Under Extreme Pressure}
% If the full title of your paper is short enough to also fit in the running head, you can omit the abbreviated paper title here. You can check as follows: if you comment out the \titlerunning line, something will appear in the header of all odd-numbered pages of your PDF from page 3 onward. This something is either the full title (in which case all is well), or the error message "Title Suppressed Due to Excessive Length". If this error message appears, you're going to want to provide an abbreviated title within the \titlerunning command, because if you won't do it, Springer will do it for you.

%N.B.: Author information (both in the \author{} and \authorrunning{} command) should only be present in the Camera-Ready Version of your paper. The version that you initially submit for review, ought to be double-blind. So, when initially submitting your paper, use:
%\author{Author information scrubbed for double-blind reviewing}
\author{Ashish Kumar\inst{1} \orcidID{0000-0002-1786-760X} \and
Durga Toshniwal \inst{1} \orcidID{0000-0002-7960-4127}  \corr
}
% You may leave out the orcidID information, if you want to.
% Use \corr to indicate the corresponding author. Note the spacing around the \corr command. Only one author can be the corresponding author.

%N.B.: comment out the \authorrunning{} command for the double-blind version of your paper submitted for review. Later, if your paper is accepted, use the command for the Camera-Ready Version.
\authorrunning{A. Kumar and D. Toshniwal }
% If there are two authors, write 'A.L. Benjamin and C.C. Broadus Jr.'
% If there are more than two authors, '[...] et al.' is used.

\institute{Indian Institute of Technology Roorkee, Roorkee, India \email{\{ashish\_k,durga.toshniwal\}@cs.iitr.ac.in}
}

\tocauthor{Ashish~Kumar, Durga~Toshniwal}
\toctitle{Modeling Text-Label Alignment for Hierarchical Text Classification}

\maketitle              % typeset the header of the contribution

\begin{abstract}
      Hierarchical Text Classification (HTC) aims to categorize text data based on a structured label hierarchy, resulting in predicted labels forming a sub-hierarchy tree. The semantics of the text should align with the semantics of the labels in this sub-hierarchy.  With the sub-hierarchy changing for each sample, the dynamic nature of text-label alignment poses challenges for existing methods, which typically process text and labels independently. To overcome this limitation, we propose a Text-Label Alignment (TLA) loss specifically designed to model the alignment between text and labels. We obtain a set of negative labels for a given text and its positive label set. By leveraging contrastive learning, the TLA loss pulls the text closer to its positive label and pushes it away from its negative label in the embedding space. This process aligns text representations with related labels while distancing them from unrelated ones. Building upon this framework, we introduce the Hierarchical Text-Label Alignment (HTLA) model, which leverages BERT as the text encoder and GPTrans as the graph encoder and integrates text-label embeddings to generate hierarchy-aware representations. Experimental results on benchmark datasets and comparison with existing baselines demonstrate the effectiveness of HTLA for HTC.

\keywords{

\and Multi-Label Classification \and NLP \and  Representation Learning.}
\end{abstract}

\section{Introduction}

In HTC, documents are assigned labels corresponding to nodes within a label hierarchy tree \cite{Zhou2020}. It has applications across diverse domains, such as scientific text categorization \cite{Aly2019}, bioinformatics \cite{Peng2016}, and online product labeling \cite{shen2021}. However, the imbalance in label frequency, coupled with the complex hierarchical structure, makes HTC a challenging task \cite{Mao2019}.

\par 
Recent approaches to HTC employ a two-encoder framework, where a text encoder processes the input text while a graph encoder captures the label hierarchy \cite{Zhou2020,Deng2021,Chen2021,Wang2022,lse-hiagm2024}.  The hierarchy is predefined based only on parent-child relationships between labels, but there are aspects to the hierarchy beyond these static links. For instance, a text sample is associated with a subset of labels that can be considered a sub-hierarchy tree. In HTC, the semantics of the text should align with the semantics of the labels in this sub-hierarchy. Aligning the semantics of the text with the semantics of the associated labels ensures that the classification model comprehensively captures the meaning conveyed in the text and accurately assigns it to the appropriate categories within the label hierarchy. This text-label alignment is dynamic since the sub-hierarchy changes for each text sample. Furthermore, existing two-encoder frameworks overlook this alignment between them as they encode text and labels separately.

\par We propose a text-label alignment (TLA) loss to address this challenge. TLA is based on the principle of contrastive learning and is formulated along lines similar to the NT-Xent loss \cite{ntxent2020}. For TLA to be effective, it is essential to carefully construct a negative label set consisting of challenging labels that are semantically distant from the text within the hierarchical structure. A hard negative mining technique is employed to select labels that demonstrate high similarity to the text sample but are not included in the positive label set, thus serving as effective negative labels. Positive and negative pairs are formed by associating each text sample with labels from the corresponding positive and negative label sets. The TLA loss increases alignment for the positive pairs, pulling text samples and their positive labels closer in the embedding space. Simultaneously, it decreases the alignment for negative pairs, thus pushing the text and negative labels away from each other in the embedding space. By dynamically aligning text and labels to the sub-hierarchy associated with each sample, the TLA loss approach inherently adjusts to the hierarchy's depth. This adaptability simplifies implementation and ensures robust performance across datasets  with varying levels of hierarchy. Furthermore, in HTC, certain labels may be more prevalent as they are assigned to several documents, while others are linked to relatively fewer documents. This variation in label frequencies can result in label imbalance, posing challenges for model training and performance. Since TLA involves explicitly modeling text-label alignment for each positive label, regardless of its frequency, it also helps mitigate the label imbalance issue.

\par Building on this, we introduce the Hierarchical Text-Label Alignment (HTLA) model, which utilizes text-label alignment for HTC. HTLA uses BERT as its text encoder and a custom implementation of GPTrans as its graph encoder. GPTrans \cite{GPTrans2024} uses transformer blocks and outperforms state-of-the-art graph models on several graph learning tasks. Its ability to model the graph from multiple dimensions makes it easily customizable for the HTC task. Within this framework, the text and label features are combined through addition, yielding a composite representation. HTLA is jointly optimized using the binary cross entropy (BCE) and TLA loss. Including TLA loss contributes to performance enhancement across datasets with simple and complex hierarchies. It models the dynamic alignment between text and labels within the hierarchical structure, addressing a challenge inadequately tackled in existing two-encoder frameworks. We summarize the contribution of our work as follows:
 \begin{itemize}
 
     \item  We propose using the Text-Label Alignment (TLA), a loss function designed to align text with its related labels in the hierarchy.
     \item  We introduce HTLA, a model that utilizes BERT as the text encoder and GPTrans as the graph encoder, optimized with BCE and TLA loss functions.
     \item  Experimental results across several datasets demonstrate the superiority of HTLA in improving classification performance. %Code is available at : \url{https://github.com/havelhakimi/TLA} 
 \end{itemize}

\section{Related Work}

HTC's existing methods can be divided into local and global approaches based on how they utilize hierarchical information. Local approaches use multiple classifiers \cite{capsule2023,huang2019,dumais2000} to make independent predictions at each node of the hierarchy, considering the local context and relationships within that specific node and its neighborhood. Global approaches model the entire hierarchical structure with a single classifier to generate predictions. Early global approaches aimed to merge the hierarchical label space using meta-learning \cite{wu2019}, recursive regularization \cite{gopal2013}, and reinforcement learning \cite{Mao2019}. These methods primarily focused on refining decoders based on hierarchical paths. The typical approach in recent studies involves enhancing flat predictions by using a graph encoder to comprehensively model the entire label structure. In their study, Zhou et al. \cite{Zhou2020} developed a graph encoder that effectively integrates existing knowledge of the hierarchical label space to acquire representations of the labels. Building upon this research, several subsequent models have emerged to explore how the hierarchical structure interacts with the text. For instance, in \cite{Chen2020}, the authors performed a joint embedding of text and labels within the hyperbolic space. Similarly, Chen et al. \cite{Chen2021} treated the problem as semantic matching, utilizing a shared space to learn representations of both text and labels. Deng et al. \cite{Deng2021} introduced an information maximization module that enhances the interaction between text and labels while imposing constraints on label representation. Zhao et al. \cite{Zhao2021} presented a self-adaptive fusion strategy capable of extracting representations from text and labels. Wang et al. \cite{Wang2022} utilized contrastive learning techniques to incorporate hierarchical information into the text encoder embedding directly. Ning et al. \cite{umpmg2023} utilizes a unidirectional message-passing mechanism to improve hierarchical label information and propose a generative model for HTC. Liu et al. \cite{lse-hiagm2024} enhance label features by introducing density coefficients for label importance in the hierarchy tree and address label imbalance with a rebalanced loss. Existing methods have employed various intricate approaches to learn hierarchical relationships and merge text-label features. However they have not emphasized on learning text-label alignment within the hierarchy. HTLA explicitly models for this dynamic alignment, ensuring that the semantics of the text align with associated labels in each sample's sub-hierarchy. This simplifies merging text and label features, requiring only addition for obtaining the composite features.

\section{Methodology}

The overall architecture of HTLA is depicted in Figure 1. This section details the components of our HTLA model, which includes the text encoder, graph encoder, generation of composite representation, and the loss functions used.

\begin{figure}
\centering
\includegraphics[width= 0.85  \textwidth]{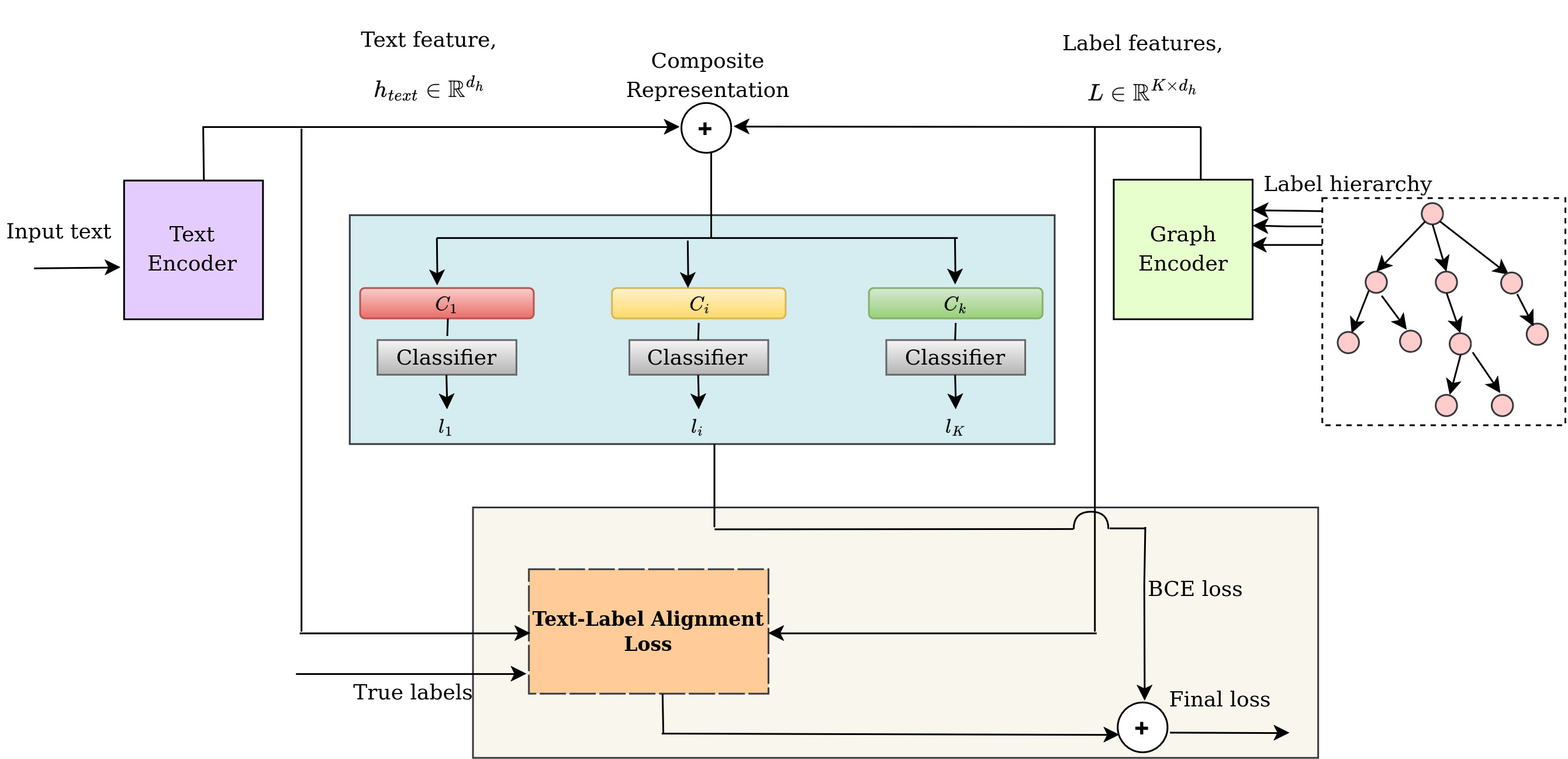}
%\captionsetup{font=scriptsize}
\caption{Architecture of the Hierarchical  Text-Label Alignment (HTLA) model. For a label $i$, its feature is combined with the text feature $h_{text}$ through addition to produce the composite feature $C_i \in \mathbb{R}^{d_h}$ for each label. A shared classifier is then utilized for each $C_i$, and the corresponding logit $l_i$ is selected from the output vector. The model is jointly optimized for BCE and TLA loss.}
\label{model}
\end{figure}

\subsection{Text Encoder}

We use BERT \cite{Devlin2019}, a transformer-based model that generates highly contextualized text embeddings by leveraging bidirectional context and pre-trained knowledge, as our text encoder. The input text is padded with two special tokens to mark the start and end of the text, as $w = \{ [CLS], w_1, w_2, \ldots, w_{n-2}, [SEP] \}$. This is then fed to the BERT encoder to produce token representations as:

\begin{equation}
%\scriptsize
%\small
H = \phi_{BERT}(w)
\end{equation}

where $ H \in \mathbb{R}^{n \times d_h} $ contains encoded representations for all $ n $ tokens. The token representation for [CLS] is chosen as the text feature for the entire sequence because it captures its contextual information, denoted as $h_{text} \in \mathbb{R}^{d_h}$.

\subsection{Graph Encoder} \label{graphencoder}

GPTrans, a graph neural network, introduces the Graph Propagation Attention (GPA) mechanism into the Transformer architecture. Unlike existing Transformer-based models that often fuse node and edge information without explicit consideration, GPA in GPTrans dynamically propagates information among nodes and edges, offering a more comprehensive and nuanced understanding of the graph structure.

\par Our customised implementation of GPTrans consists of three main components: Feature Initialization, GPA, and \textit{LabelEnhancer} module
\subsubsection{Feature Initialization} The node and edge features are initialized in this component. For each label node $i$, the node feature $g_i \in \mathbb{R}^{d_h}$ is initialized  as:
\begin{equation}
%\scriptsize
%\small
 g_i =embed_{node}(i)+ embed_{name}(i) \label{node_feat}   
\end{equation}

\begin{itemize}
    \item $embed_{node}(.)$  is a  learnable embedding function that generates embedding of size $d_h$ for each input node to capture essential node characteristics. 
    \item $embed_{name}(.)$ function uses the BERT tokenizer to tokenize each label name, calculates the average of the token embeddings, and assigns it to the label. This process aids in extracting semantic information and summarizing distinctive characteristics associated with each label. The weights used for learning text embeddings with BERT are shared with $embed_{name}(.)$, ensuring informativeness in label features.
\end{itemize}

The edge feature $x_{ij} \in \mathbb{R}^{d_p} $ for each pair of nodes is initialized as:
\begin{equation}
%\scriptsize
%\small
x_{ij}= S_{f(i, j)}+ E_{ij}  \label{edge_feat}
\end{equation}
\begin{itemize}
    \item $S_{f(i, j)}$ is the  spatial encoding component, indexed by distance measure function $f(i, j)$, representing the distance between nodes $i$ and $j$. It is a learnable embedding of size $d_p$.

    \item $E_{ij}$ is the edge encoding component, accounting for edge weights along the unique path $(e_1, e_2, ..., e_D)$ connecting nodes $i$ and $j$ in the label hierarchy tree, where $D=f(i,j)$. The computation for $E_{ij}$ involves averaging the edge weights along this path, expressed as $\frac{1}{D} \sum_{z=1}^{D} w_{e_{z}}$, where each $w_{e_z} \in \mathbb{R}^{d_p}$ represents the weight parameter for the corresponding edge $e_z$.
\end{itemize}

Finally, matrices $g \in \mathbb{R}^{K \times d_h}$ and $x \in \mathbb{R}^{K \times K \times d_p}$ are formed by stacking node and edge features, respectively, where $K$ is the number of label nodes.

\subsubsection{Graph Propagation Attention}
This modified attention module explicitly defines the information flow between nodes and edges, allowing for the capture of both local and higher-order relationships within the label hierarchy. To simplify, we assume single-head self-attention in the following equations.

\par In the \textbf{node-to-node flow},  self-attention is improved by incorporating edge information. For this edge features $x$ are transformed using   $W_1 \in \mathbb{R}^{d_p \times n_{head}}$ which is then added to the attention map. The update node features $g' \in \mathbb{R}^{K \times d_h}$ are then computed by multiplying with value matrix $V$ as:

\begin{equation}
%\scriptsize
%\footnotesize
%\small
\begin{gathered}
	x' = xW_1; \;  
	A = \frac{(gW_Q)(gW_K)^T}{\sqrt{dim_h}} + x'; \;  
	g' = softmax(A)V
\end{gathered}
\end{equation}

where $W_Q, W_K, W_V \in \mathbb{R}^{d_h \times d_h}$, $V=gW_V$, and $dim_h = d_h/n_{head}$ refers to the size of each head.

\par The \textbf{node-to-edge} flow updates the edge features based on attention patterns observed during node-to-node interactions. The attention scores $A \in \mathbb{R}^{K \times K\times n_{head}}$ are combined with their softmax values, creating a weighted sum, which is then transformed by the matrix $W_2 \in \mathbb{R}^{n_{head} \times d_p}$ as:

\begin{equation}
%\scriptsize
%\small
 x'=(A+ softmax(A))W_2
\end{equation}

In the \textbf{edge-to-node} flow, weights are computed based on edge features $x' \in \mathbb{R}^{K\times K \times d_p}$ calculated in previous step. Subsequently, a weighted sum of edge features is utilized to update node features, followed by linear transformations $W_3 \in \mathbb{R}^{d_p \times d_h}$ and $W_4 \in \mathbb{R}^{d_h \times d_h} $, as: 
\begin{equation}
%\scriptsize
%\small
\begin{gathered}
	g''=(sum(x'.softmax(x'), dim=1))W_3;\;
	g'''=(g'+g'')W_4
\end{gathered}
\end{equation}

For more details on GPA please refer to the original paper \cite{GPTrans2024}.

\subsubsection{LabelEnhancer}
The label node features $g''' \in \mathbb{R}^{K \times d_h}$ generated by GPA serve as input to $LabelEnhancer$, a multi-layered neural network. It refines these node representations, producing the final label features $L \in \mathbb{R}^{K \times d_h}$ as:
 \begin{equation}
 %\small
     L=LabelEnhancer(g''')
 \end{equation}

\subsection{Generation of Composite Representation}

To create a composite representation, we merge the text and label features by adding them together. In the label feature matrix \(L \in \mathbb{R}^{K \times d_h}\), each \( f_i\) represents the feature vector for label \( i\). We enhance the label feature \( f_i \) by incorporating the text feature \( h_{\text{text}} \in \mathbb{R}^{d_h} \) from the corresponding sample. This  results in a composite feature \(C_i \) that captures both the textual context and the specific characteristics of label \(i\). Subsequently, this composite feature is fed into the classifier. The logit score \(l_i\) for label \(i\) is calculated as the \(i^{th}\) element of the resulting classifier output vector, and the predicted output for label $i$ is obtained after applying $sigmoid(.)$
on $l_i$. This process is formally defined in Equation  \ref{fusion_labeltext} below:

\begin{equation}
%\small
%\scriptsize
\begin{gathered}
C_i = h_{text} + f_i ; \quad l_i = (W_c^T C_i + b)_i; \quad \hat{y_i}=sigmoid(l_i) \label{fusion_labeltext} 
\end{gathered}
\end{equation}

where $W_c \in \mathbb{R}^{d_h \times K}$ and $b \in \mathbb{R}^{K}$ are weights and bias of the classifier. The parameters of the classifier ($W_c$ and $b$) are shared across all labels, ensuring consistency in predictions.

\subsection{Loss Functions}

\subsubsection{Text-Label Alignment Loss}

%L_{tl} = \frac{1}{N_b} \sum_{i=1}^{N_b} \frac{1}{\lvert P_1(i) \rvert} \sum_{p \in P_1(i)} \left( -\log \exp \left( \frac{(z_i, %e_p)}{\tau} \right) \right) \times \sum_{a \in A_1(i)} \exp \left( \frac{(z_i, e_a)}{\tau} \right)

%\begin{equation}
%    L_{tla}=\frac{1}{M} \sum_{i=1}^{M}  \frac{1}{\lvert P(i) \rvert} \sum_{p \in P(i)} - \log \frac{\exp(sim(h_{text_{i}},e_p)/ %%\tau)}{\sum_{s \in S} \exp(sim(h_{text_{i}},e_s) / \tau)
%}
%\end{equation}

%In HTC, the fundamental challenge lies in establishing meaningful semantic associations between text samples and their %corresponding labels within a hierarchical structure.

In HTC, it is desired that the representation of a sample not only reflects its semantic content but also aligns closely with its positive labels while remaining distinct from negative labels in the embedding space. The challenge lies in identifying negative labels to establish the necessary contrasting relationship for alignment. We use hard negative mining to select a set of negative labels for each sample. Once both positive and negative labels are identified, we form pairs with the text samples and compute the TLA loss. This encourages closer alignment between text and its positive labels while maximizing dissimilarity with negative ones.

\par The TLA loss operates on a batch of text samples, denoted as \( M \), each associated with a set of positive labels \( P(i) \), where \( i \) represents the index of the text sample. For each sample, we obtain a set of negative labels with high similarity scores to the text sample, excluding those already identified as positive labels and denote it as \( N(i) \). A positive pair is formed consisting of \( (h_{text_{i}}, f_p) \) where \( f_p \) denotes the label feature for label \( p \in P(i) \) and \( h_{text_{i}}  \) represents text feature of the \(i^{th}\) sample. Similarly, a negative pair is formed consisting of \( (h_{text_{i}}, f_n) \), where \( f_n \) denotes the label feature for label \( n \in N(i) \). The TLA loss is then defined as:

\begin{equation} 
%\scriptsize
%\small
Loss_{TLA} = \frac{1}{M} \sum_{i=1}^{M} \frac{1}{\lvert P(i) \rvert} \sum_{p \in P(i)} - \log \left( \frac{\exp(\text{sim}(h_{text_{i}},f_p)/\tau)}{\sum_{s \in S(i)} \exp(\text{sim}(h_{text_{i}},f_s) / \tau)} \right)  \label{TLA}
\end{equation}

where \( \text{sim(.)} \) computes cosine similarity, \( \lvert P(i) \rvert \) denotes cardinality of label set \( P(i)\), \( S(i) = N(i) \cup P(i) \), and \( \tau \in \mathbb{R}^{+} \) controls temperature. Algorithm \ref{TLAalgo} outlines the steps to compute TLA loss for a batch of text samples.

\begin{algorithm}
\caption{Text-label alignment (TLA) loss}
\label{TLAalgo}
%\scriptsize
%\small
%\tiny
\begin{algorithmic}[1]
\State \textbf{Input:} Text features \( Z (M \times d_h) \), Label features \( L (K \times d_h) \), True labels \( Y ( M \times K) \), Temperature \( \tau \)
\State \textbf{Output:} TLA loss, \( Loss_{TLA}\)
\State  \( P \gets \{\} \), \( N \gets \{\} \)   \Comment{\textit{  Initialize set for pos and neg labels}}
\State  \( sim\_mat \gets  cos\_sim(Z, L^{T}) \) \Comment{\textit{ Compute cosine similarity }}

 \State  \( P \gets  \{ p_i \mid p_i = \{j \mid Y_{ij} = 1\}, \forall i \in \{1, 2, \ldots, M\} \} \) 
\Comment{\textit{  Add indices of positive labels }}

\For{\textbf{each} \( i \) \textbf{from} \( 1 \) \textbf{to} \( M \)}  \Comment{\textit{HardMining to get neg label set }}
    \State \( N[i] \gets \{\} \)
    \State \( p\_labels \gets P[i] \)
    \State \( neg\_sim \gets sim\_mat[i] \)
    \For{\textbf{each} \( label \) \textbf{in} \( p\_labels \)}
        \State \( neg\_sim[label] \gets -\infty \) \Comment{\textit{Set similarity to neg infinity for pos labels}}
    \EndFor
    \State \( sorted\_indices \gets argsort(neg\_sim, descending=True) \)
    \State \( hard\_negative\_labels \gets \{ sorted\_indices[k] \mid k \in [1, len(p\_labels)] \} \)
    \State  \( N[i] \gets  N[i] \cup   hard\_negative\_labels  \)
\EndFor

\State $S \gets  \{\}$

\For{\textbf{each} $i$ \textbf{from} $1$ \textbf{to} $M$} \Comment{\textit{Combine pos and neg label sets}}

    \State $S[i] \gets P[i] \cup N[i]$
\EndFor
\State Compute \( Loss_{TLA}\)  using Equation \ref{TLA}

\State \textbf{return} \( Loss_{TLA}\)
\end{algorithmic}
\end{algorithm}

\subsubsection{Binary Cross Entropy Loss}
While TLA enhances semantic alignment by aligning text with its labels, BCE complements this by emphasizing the correctness of label predictions, enabling the model to learn the distinctive features of each label independently. BCE loss for a batch of \( M\) samples is formulated as:

\begin{equation}
%\small
%\scriptsize
Loss_{BCE} = - \frac{1}{M} \sum_{i=1}^{M} \sum_{j=1}^{K} \left( Y_{ij} \log(\hat{Y}_{ij}) + (1 - Y_{ij}) \log(1 - \hat{Y}_{ij}) \right)
  \label{bce}
\end{equation} 
where $Y \in \mathbb R^{M \times K}$ represents the true label values and $\hat{Y} \in  \mathbb R^{M \times K}$ represents the predicted label probabilities.
\subsubsection{Final Loss}

The final loss for the HTLA model is obtained by the sum of both BCE and TLA losses as:
\begin{equation}
%\small
%\scriptsize
    Loss_{HTLA}=Loss_{BCE}+Loss_{TLA}
\end{equation}

\section{Experiments}

\subsection{Datasets and Evaluation Metrics}

\begin{table}[t]
    \centering
   % \scriptsize
    %\tiny
    %\captionsetup{font=scriptsize}
    \caption{Statistical details for the WOS, RCV1-V2, and NYT datasets. $|Level|$ indicates the number of hierarchy levels, $|L|$ is the total label count, and Mean-$|L|$ denotes the mean number of labels per sample}
    \label{tab1}
    \renewcommand {\arraystretch}{1.2}
    \begin{tabular}{ccccccc} 
    \toprule
    Dataset & $|Level|$ & Train & Val & Test & $|L|$ & Mean-$|L|$ \\
    \midrule
    WOS & 2 & 30070 & 7518 & 9397 & 141 & 2.0 \\
    RCV1-V2 & 4 & 20833 & 2316 & 781265 & 103 & 3.3 \\
    NYT & 8 & 23345 & 5834 & 7292 & 166 & 7.6 \\
    \bottomrule
    \end{tabular}
\end{table}

We conducted experiments and model evaluations using three datasets: WOS \cite{Kowsari2017}, RCV1-V2 \cite{Lewis2004}, and NYT \cite{Sandhaus2008}.  The WOS dataset contains abstracts from scientific papers, with their corresponding labels arranged in a single-path hierarchy. RCV1-V2 and NYT are news categorization datasets with multiple label paths in the hierarchy.
Table \ref{tab1} provides detailed statistics for each dataset. In line with previous HTC studies \cite{Chen2021,Deng2021,Wang2022,lse-hiagm2024}, we followed the label hierarchy taxonomy, data preprocessing steps and train-val-test splits outlined in \cite{Zhou2020}.
We evaluated performance using the Micro-F1 and Macro-F1 scores, consistent with previous research \cite{Zhou2020,Chen2021,Deng2021,Wang2022,lse-hiagm2024}.

\subsection{Implementation Details}

In our implementation, we use the \textit{bert-base-uncased} model from the hugging face transformers library \cite{hugggingfacetransformers} as our BERT-based text encoder. We utilize a single layer of the GPTrans block, which includes a multi-headed attention mechanism with 12 attention heads ($n_{head}$). The edge feature size, $d_p$, is set to 30 for all datasets, determined through grid search on validation set. As for the node feature size, $d_h$, we keep it identical to the text representation size of 768. The temperature hyperparameter $\tau$ for TLA is set to 0.07 for all datasets.  During training, we use a batch size of 10 and opt for the Adam optimizer with a learning rate of 1e-5. Our model is implemented in PyTorch and trained end-to-end. We assess the model's performance on the validation set after each epoch and halt the training procedure if the Macro-F1 score does not show improvement for six consecutive epochs. The architectural details of the $LabelEnhancer$ module are outlined in Table \ref{lbl_ref}.

\begin{table}[]
  \renewcommand{\arraystretch}{1.2}
  %\scriptsize
  %\tiny
  \centering
    %\captionsetup{font=scriptsize}
    \caption{Layer specification for the $LabelEnhancer$ module}   \label{lbl_ref}

  \begin{tabular}{ll}
    \hline
    \textbf{Layer} & \textbf{Input/Output Shape} \\
    \hline
    Input  & $K \times d_h$  (\textit{label features} $g'''$) \\
    %Layer Normalization & $K \times d_h$ \\
    Linear & $K \times d_h / K \times 4d_h$ \\
    Activation (GELU) & $K \times 4d_h / K \times 4d_h$ \\
    Dropout  & $K \times 4d_h / K \times 4d_h$ \\
    Linear & $K \times 4d_h / K \times d_h$  \\
    Dropout  & $K \times d_h / K \times d_h$  (\textit{intermediate label features} $\hat{g}$)\\
    Residual Connection   & $K \times d_h / K \times d_h$ ($g'''+\hat{g}$) \\
    Layer Normalization & $K \times d_h / K \times d_h$  (\textit{Final label features} $L$) \\
    \hline
  \end{tabular}

\end{table}

\subsection{Experimental results}

\begin{table*}[t]
\renewcommand{\arraystretch}{1.05}
\centering

%\scriptsize
%\tiny
%\captionsetup{font=scriptsize}
\caption{
Comparison of results across three datasets. We report average score of 8 random runs for our implemented models(denoted with an asterisk (*)), with the second best results among our implemented models underlined. Results for other models were sourced from their respective papers.} \label{tab:model-performance}
\begin{tabularx}{1.0 \textwidth}{p{3.7cm}XXXXXX} %{p{2.5cm}p{1cm}p{1cm}p{1cm}p{1cm}p{1cm}p{1cm}} % {p{3.2cm}XXXXXX}
\toprule
\multirow{2}{*}{Model} & \multicolumn{2}{c}{WOS } & \multicolumn{2}{c}{RCV1-V2} & \multicolumn{2}{c}{NYT} \\ \cmidrule{2-7}
                       & MiF1      & MaF1      & MiF1      & MaF1      & MiF1      & MaF1      \\ \midrule
%\multicolumn{7}{c}{\textbf{Global Models}} \\ \midrule

TextCNN \cite{Zhou2020}  & 82.00 & 76.18  & 79.37 & 59.54 & 70.11 & 56.84 \\
TextRCNN \cite{Zhou2020} & 83.55 & 76.99  & 81.57 & 59.25 & 70.83 & 56.18 \\
HiLap-RL \cite{Mao2019}  & - & -       & 83.30 & 60.10 & 74.60 & 51.60 \\
HiAGM \cite{Zhou2020} & 85.82 & 80.28 & 83.96 & 63.35 & 74.97 & 60.83 \\
HTCInfoMax \cite{Deng2021} & 85.58 & 80.05 & 83.51 & 62.71 & - & - \\
HiMatch  \cite{Chen2021} & 86.20 & 80.53 & 84.73 & 64.11 & - & - \\
LSE-HiAGM \cite{lse-hiagm2024} & 86.01 & 80.01 & 83.86 & 64.57 & 75.01 & 61.29 \\
\midrule
%\multicolumn{7}{c}{\textbf{Pretrained Language Models}} \\ \midrule
BERT+HiAGM \cite{Wang2022} & 86.04 & 80.19 & 85.58 & 67.93 & 78.64 & 66.76 \\
BERT+HTCInfoMax \cite{Wang2022} & 86.30 & 79.97 & 85.53 & 67.09 & 78.75 & 67.31 \\
HiMatch-BERT \cite{Chen2021} & 86.70 & 81.06 & 86.33 & 68.66 & - & - \\
HGCLR \cite{Wang2022} & 87.11 & 81.20 & 86.49 & 68.31 & 78.86 & 67.96 \\
BERT\textsuperscript{*} & 85.85 & 79.93 & 86.14 & 67.10 & 78.65  & 66.31 \\
BERT-GPTrans\textsuperscript{*} & 86.74 & 80.62 & \underline{86.28} & \underline{68.19} & \underline{78.89} & \underline{67.34} \\ 
HGCLR\textsuperscript{*} & \underline{87.09} & \underline{81.08} & 86.27 & 68.09 & 78.53 & 67.20 \\ 
HTLA\textsuperscript{*} & \textbf{87.38} & \textbf{81.88} & \textbf{87.14} & \textbf{70.05} & \textbf{80.30} & \textbf{69.74} \\
 \bottomrule
\end{tabularx}

\end{table*}

Table \ref{tab:model-performance} displays the results of HTLA and compares them with various baselines. For a detailed analysis and comparison, we also implemented fine-tuned BERT (\textit{bert-base-uncased} from Hugging Face) and the BERT-GPTrans and HGCLR\cite{Wang2022} alongside HTLA. While BERT employs a flat multi-label classification without considering hierarchy, BERT-GPTrans models hierarchy and is trained solely on the BCE loss. HGCLR uses contrastive learning to embed hierarchy information into BERT encoder.
HGCLR,  constructs positive samples for input text by masking unimportant tokens from the representation obtained through cross-attention between text and label features. The masking of tokens is determined by a threshold value, an additional hyperparameter that needs tuning for each dataset. This can inevitably introduce noise and overlook label correlations if the threshold is not appropriate. HTLA aligns text with its positive labels on a per-sample basis, ensuring that relationships between labels within the sub-hierarchy tree are implicitly captured. We conducted a one-sided paired t-test with significance level set at 0.05 to determine whether HTLA yield significantly improved outcomes. t-tests are recommended for assessing hypotheses related to average performance\cite{CUNHA2021}, and they remain robust even when normality assumptions are violated \cite{Hull1993}. Across all datasets, the performance scores of HTLA show a statistically significant improvement. Further details regarding the statistical tests can be found in Appendix \ref{stat}.

\par  For the WOS, RCV1-V2 and NYT datasets, the HTLA shows a 0.8\% , 1.9\%, 2.4\%, increase in the Macro-F1 (MaF1) compared to the second best. HTLA is more effective in enhancing text-label alignment for datasets with deeper hierarchies like RCV1-V2 and NYT, where multiple positive labels exist at each level. However, in WOS, characterized by a shallow two-level hierarchy and only one related label per level, the improvements are comparatively modest. Also, the improvements in Micro-F1(MiF1) are somewhat limited across all datasets, mainly due to its computation method. MiF1 aggregates the confusion matrix for each label, making it sensitive to predominant labels characterized by high frequencies. Conversely, MaF1 computes distinct F1 scores for each label and then averages them, assigning equal importance to all labels, irrespective of their occurrence frequency. The considerable increase in MaF1 suggests that our models effectively handle label imbalance and improve the classification of less common labels. 

\subsection{Analysis}

\subsubsection{ Performance amid label imbalance }

\begin{figure}[h]
\centering
%\scriptsize
\begin{subfigure}{0.3\textwidth}
  \includegraphics[width=\linewidth]{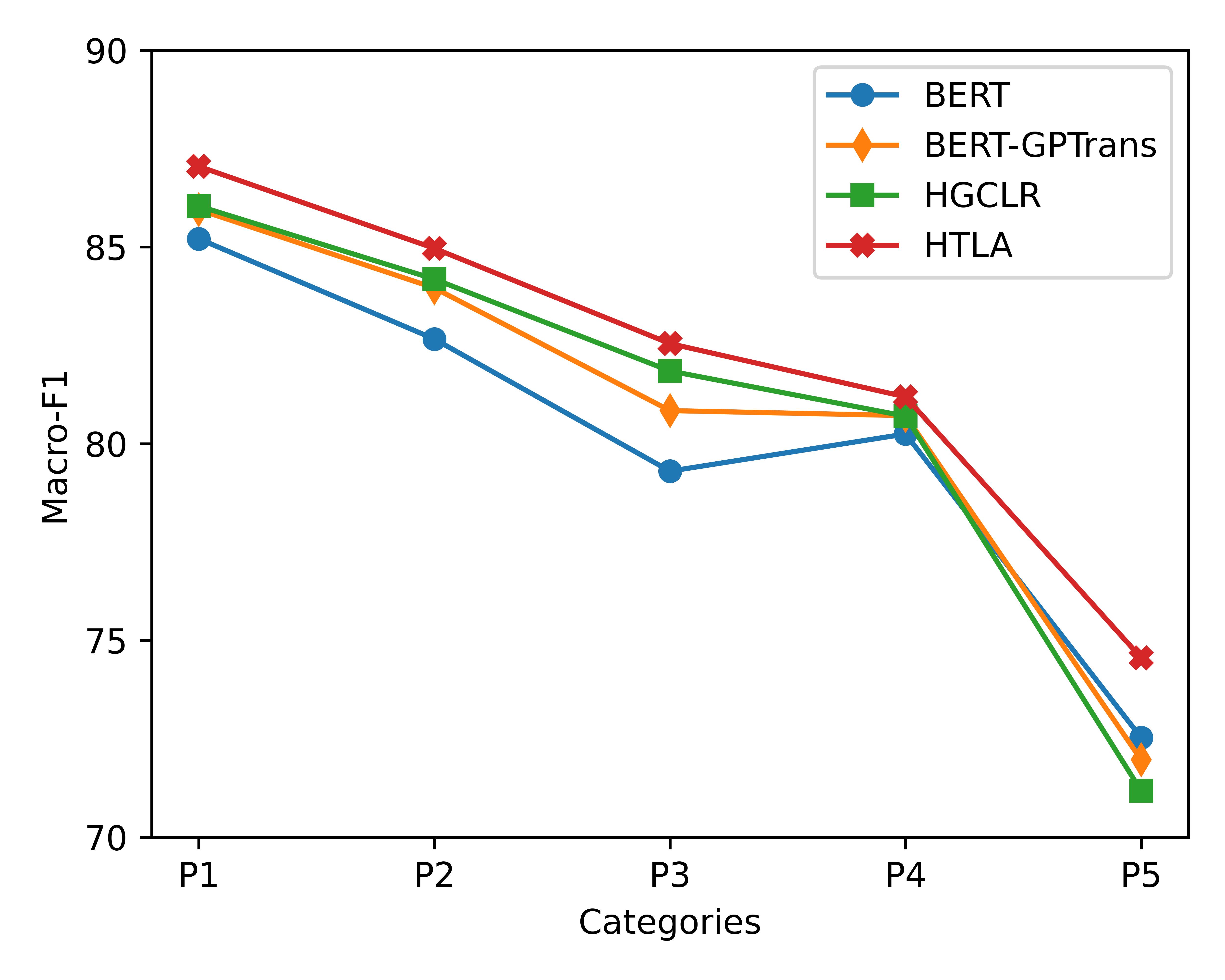}
  \captionsetup{font=scriptsize}
  \caption{WOS }\label{wos_pop}
\end{subfigure}
\begin{subfigure}{0.3\textwidth}
  \includegraphics[width=\linewidth]{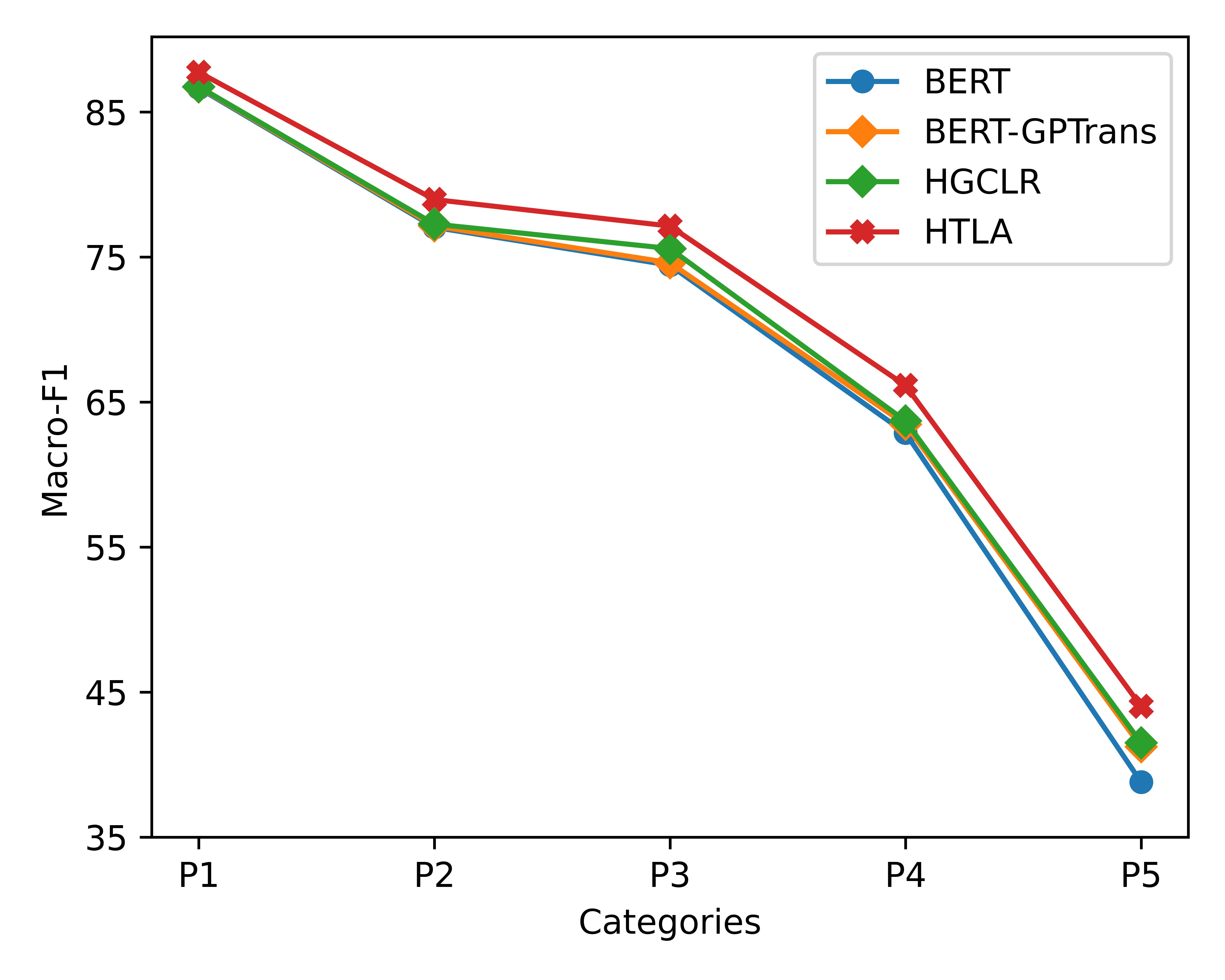}
  \captionsetup{font=scriptsize}
  \caption{RCV1-V2 }\label{rcv_pop}
\end{subfigure}
\begin{subfigure}{0.3\textwidth}
  \includegraphics[width=\linewidth]{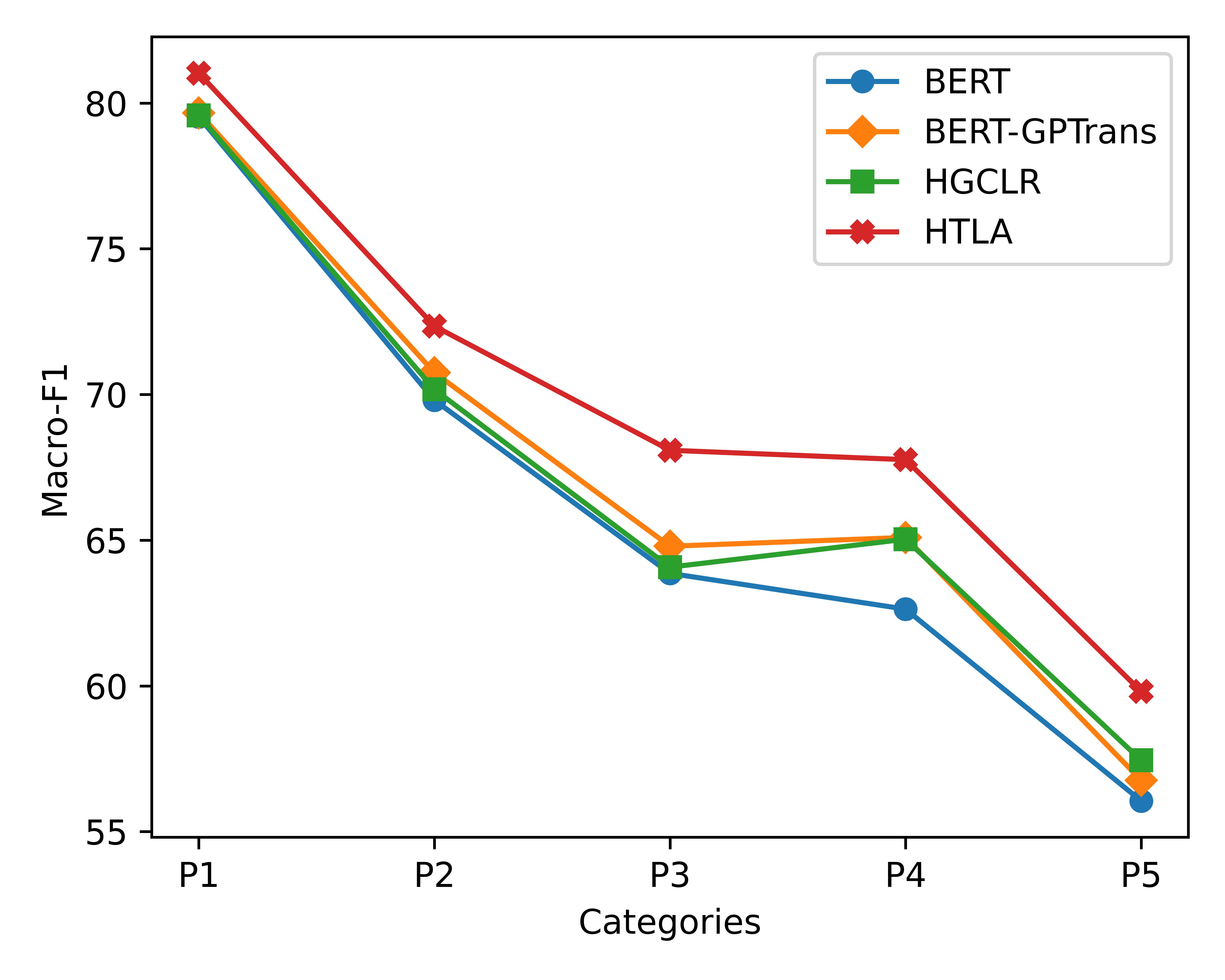}
  \captionsetup{font=scriptsize}
  \caption{NYT }\label{nyt_pop}
\end{subfigure}
%\captionsetup{font=scriptsize}
\caption{ Model performance across label prevalence categories} \label{combined_pop}
\end{figure}

Evaluating a model's performance across different levels of label prevalence can provide insight into its efficacy under label imbalance. To assess model performance, we arrange the labels in descending order by the number of associated documents and divide them into five equally sized groups, denoted P1 to P5. Each group contains 20\% of the labels, with P1 comprising the most prevalent labels and P5 the least. Figure \ref{combined_pop} illustrates performance across these prevalence categories. HTLA outperforms other models, particularly for less prevalent labels in category P5, demonstrating its effectiveness in addressing label imbalance.

\subsubsection{Performance across hierarchy levels}

\begin{figure}[h]
\centering
%\scriptsize

\begin{subfigure}{0.32\textwidth} % Adjust the width as necessary
  \centering
  \includegraphics[width=\linewidth]{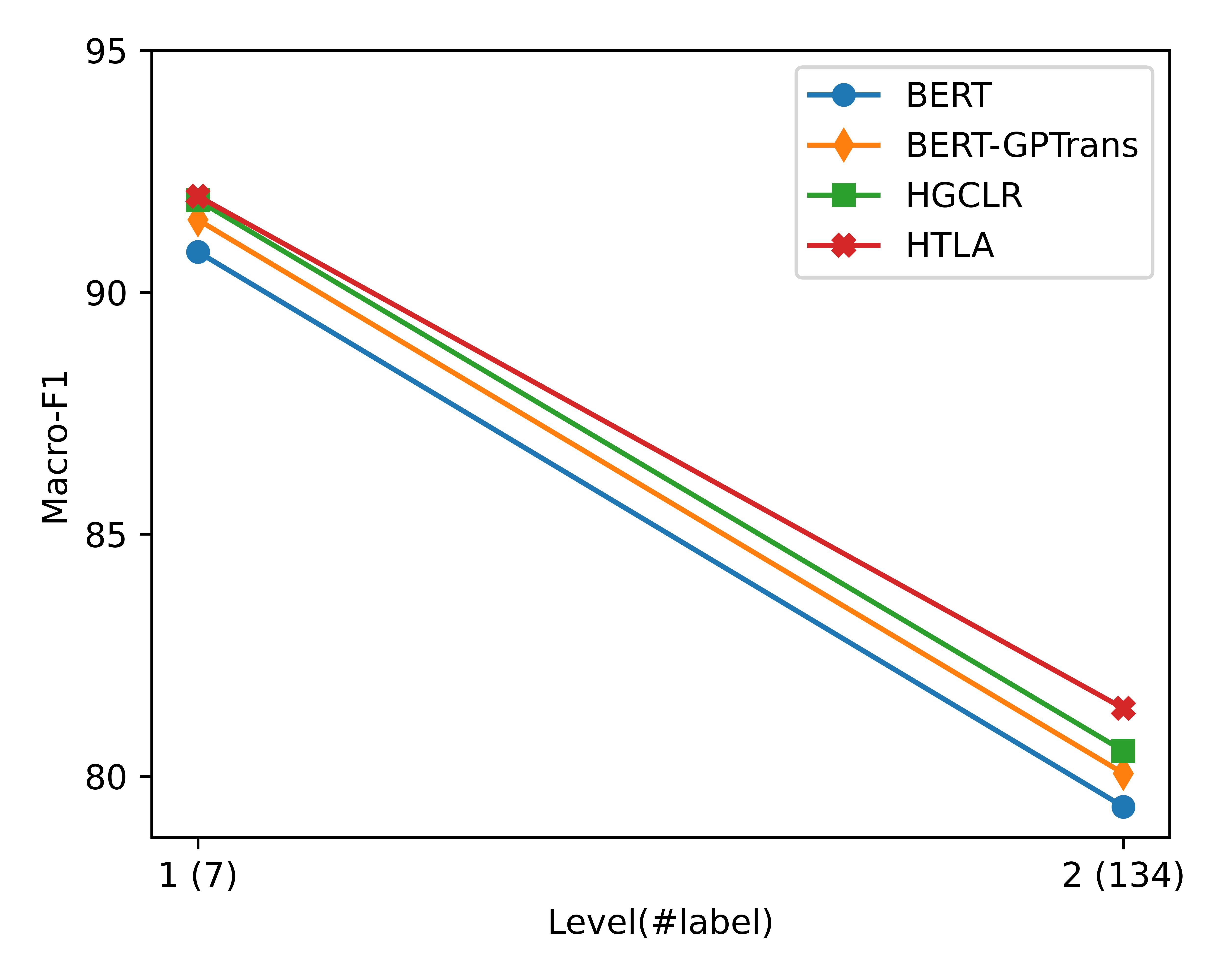}
  \captionsetup{font=scriptsize}
  \caption{WOS}\label{wos_lvlmac}
\end{subfigure}%
\begin{subfigure}{0.34\textwidth}
  \centering
  \includegraphics[width=\linewidth]{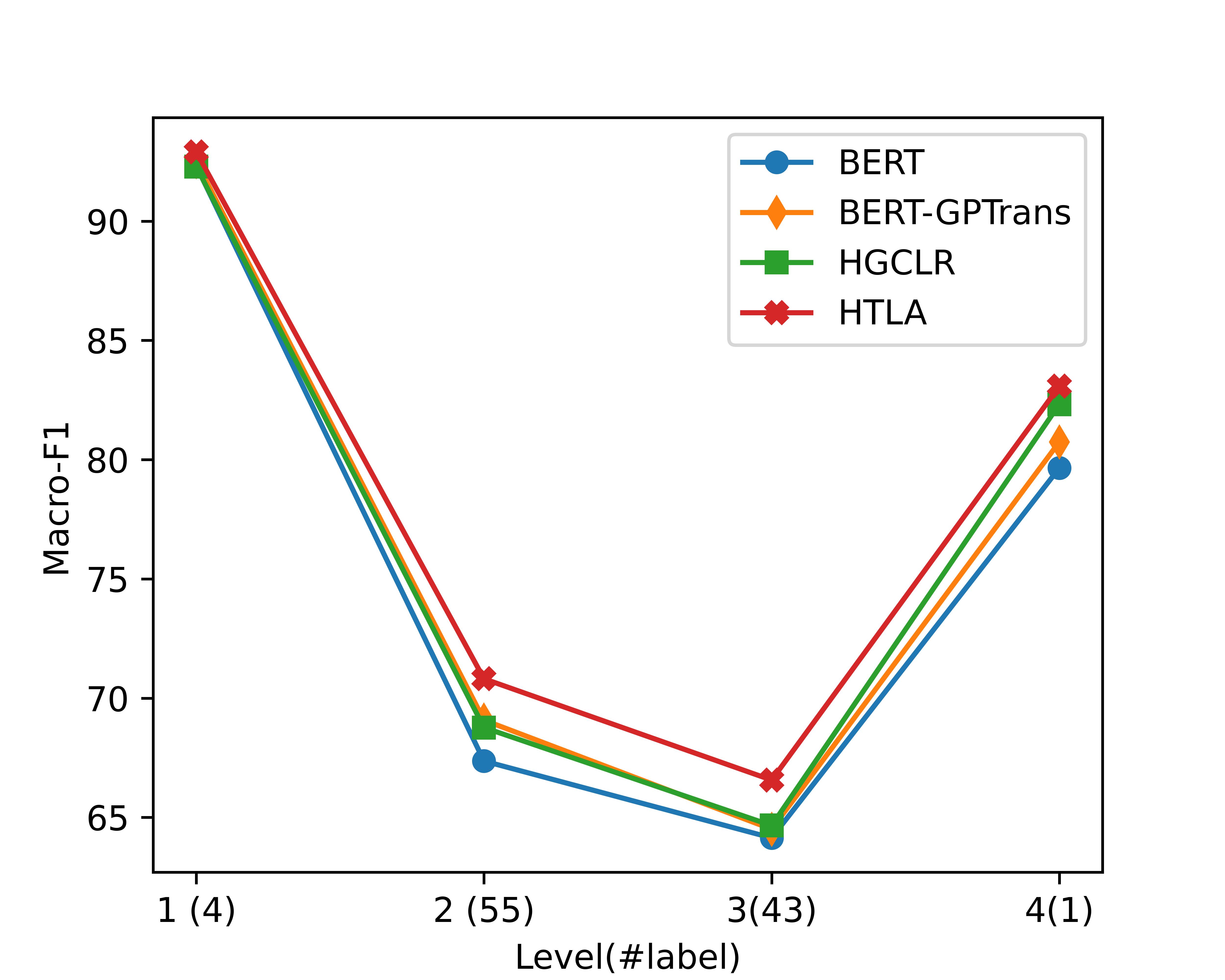}
  \captionsetup{font=scriptsize}
  \caption{RCV1-V2}\label{rcv_lvlmac}
\end{subfigure}%
\begin{subfigure}{0.32\textwidth}
  \centering
  \includegraphics[width=\linewidth]{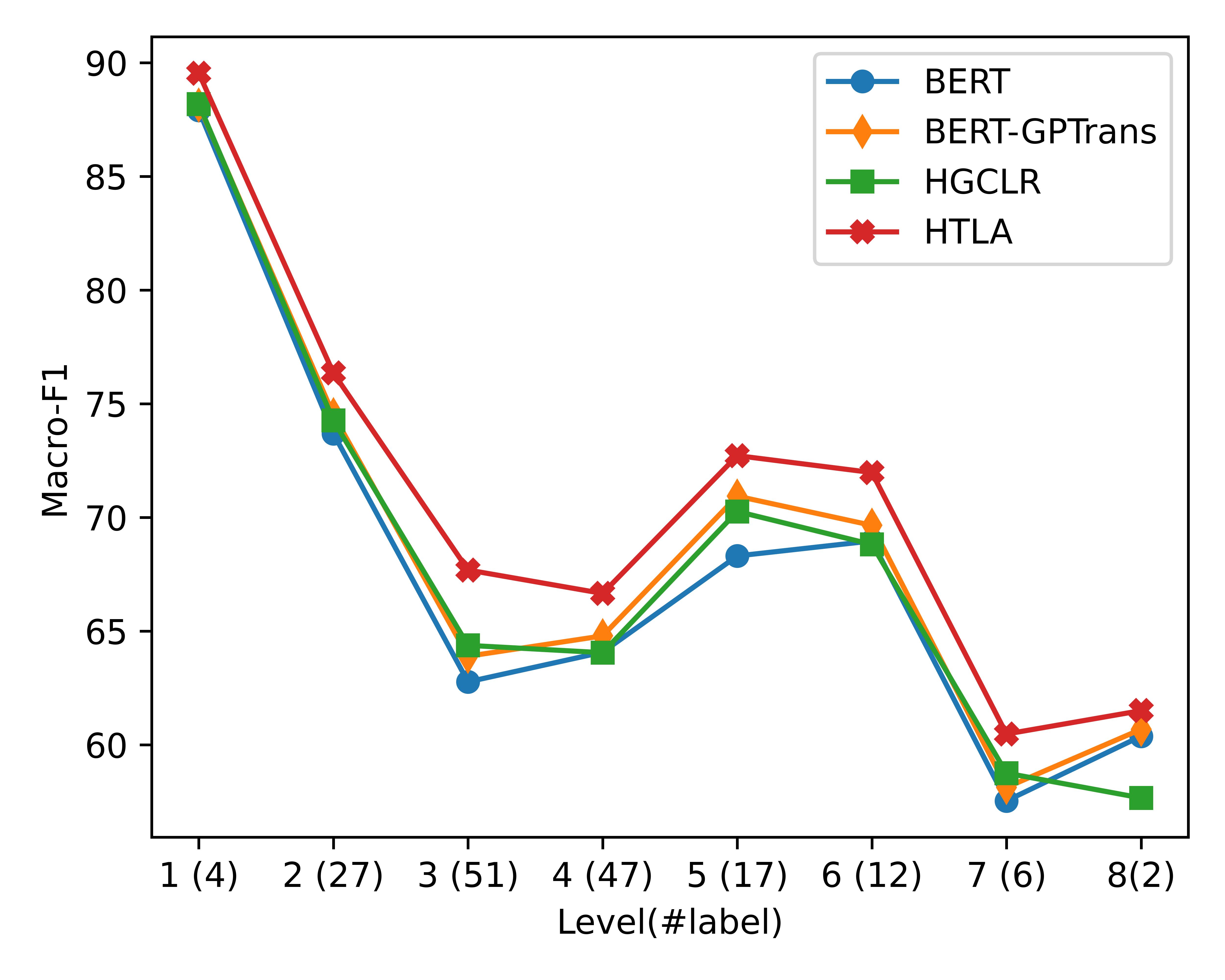}
  \captionsetup{font=scriptsize}
  \caption{NYT}\label{nyt_lvlmac}
\end{subfigure}

%\captionsetup{font=scriptsize}
\caption{Model performance across hierarchy levels}\label{combined_gran}
\end{figure}

Labels within hierarchies can span from general to highly specific categories. Models that excel at capturing broad patterns may struggle with finer distinctions, particularly at lower levels of the hierarchy. Figure \ref{combined_gran} illustrates the model performance across hierarchy levels for datasets with shallow hierarchies (WOS) and those with deeper hierarchies (RCV1-V2 and NYT). In WOS, HTLA outperforms its counterparts, particularly for fine-grained labels at the second level. In RCV1-V2, characterized by numerous ambiguous labels at the second level and fine-grained labels at levels two and three, HTLA consistently outperforms other models. In NYT, which features the deepest hierarchy and an uneven distribution of labels across different levels, HTLA exhibits superior performance, especially at the deeper levels.

\subsubsection{Performance based on the number of label paths}

\begin{figure}[h]
\centering
%\scriptsize

\begin{subfigure}{0.34\textwidth} % Adjust the width as necessary
  \centering
  \includegraphics[width=\linewidth]{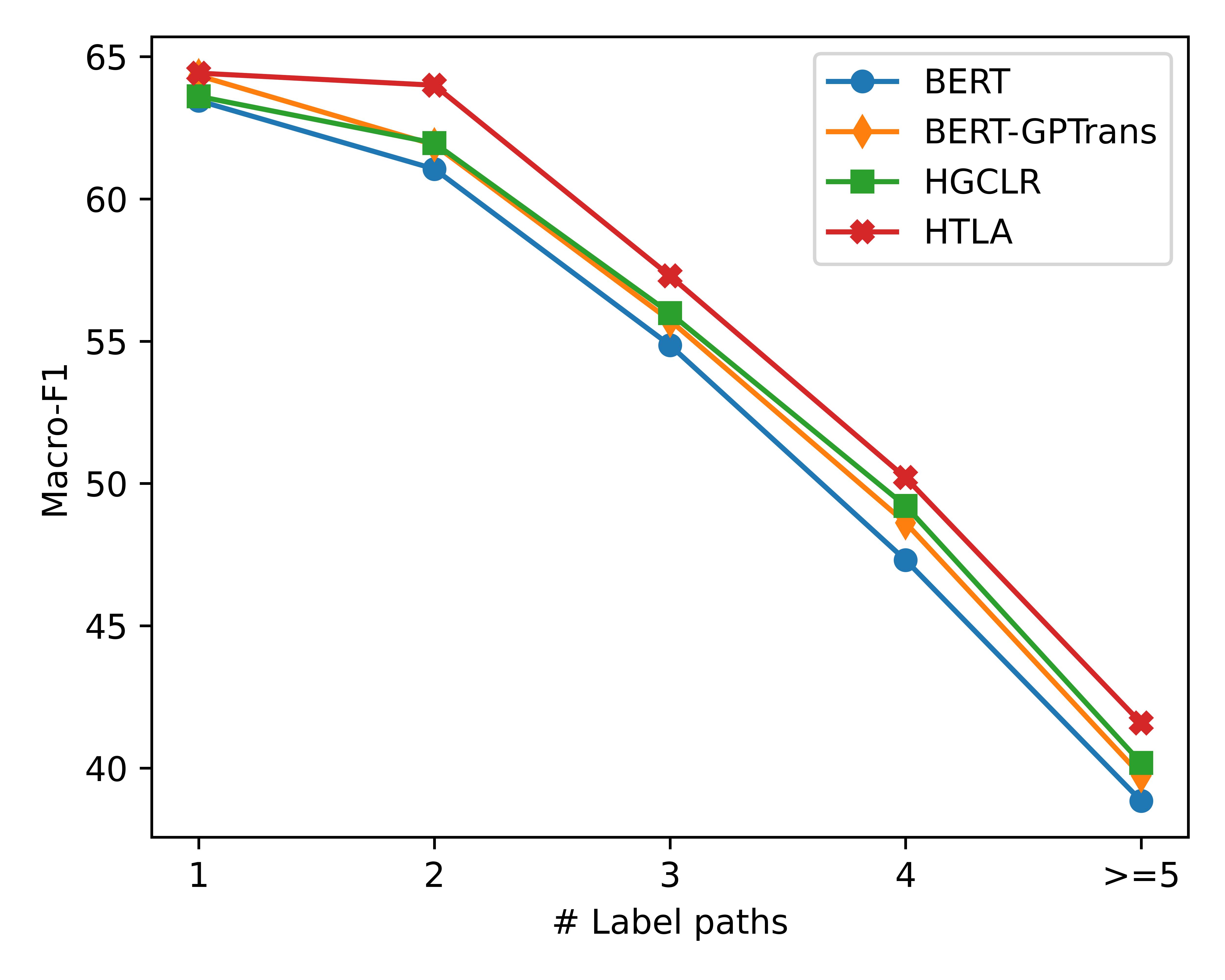}
  \captionsetup{font=scriptsize}
  \caption{RCV1-V2}\label{rcvpathmac}
\end{subfigure}%
\begin{subfigure}{0.34\textwidth}
  \centering
  \includegraphics[width=\linewidth]{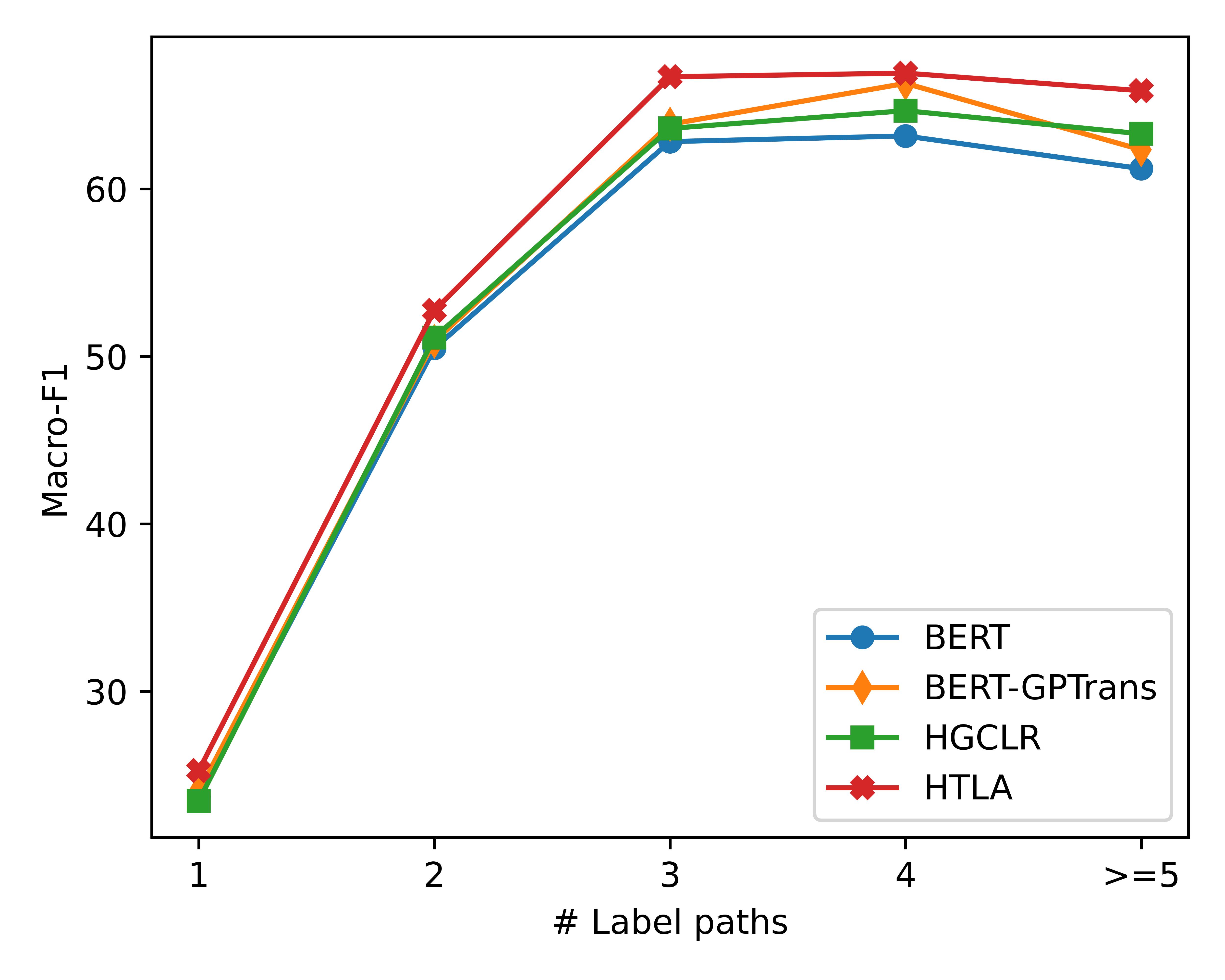}
  \captionsetup{font=scriptsize}
  \caption{NYT}\label{nytpathmac}
\end{subfigure}%

%\captionsetup{font=scriptsize}
\caption{Model performance across label paths}\label{combined_path}
\end{figure}

We conduct a performance analysis for datasets with multiple label paths by grouping samples based on the number of paths they traverse in the label hierarchy. Figure \ref{combined_path} illustrates the performance on samples for both the RCV1-V2 and NYT datasets. For both datasets, HTLA demonstrates a performance boost compared to other models as the number of label paths increases. These results indicate that HTLA excels in handling hierarchical structures with multiple label paths, making it a robust performer for datasets with intricate and complex hierarchies.

\subsubsection{Ablation Study and Model Generalizability}

Our model, HTLA, leverages TLA Loss and customized GPTrans, which consists of $embed_{node}(.)$ and $embed_{name}(.)$ functions to initialize features, along with a $LabelEnhancer (LE)$ module to refine label features. To assess each component's impact, we systematically removed them one at a time. The first part of Table \ref{ablation} presents ablation results for HTLA. The results clearly indicate that the removal of these components leads to a decrease in performance, while HTLA, with all components intact, achieves the best performance among the compared models. Furthermore, to demonstrate model generalizability, we conducted experiments on two additional text datasets: AAPD \cite{aapd2018} and BGC \cite{Aly2019}, using the same train-val-test splits as the original studies. Further details regarding these datasets are provided in Appendix \ref{bgcapd}. The second part of Table \ref{ablation} presents the results on these additional datasets, where the use of HTLA shows a performance boost compared to other models.

\begin{table}[]
\renewcommand{\arraystretch}{1.1}
\centering
%\tiny
%\scriptsize
%\captionsetup{font=scriptsize}
\caption{Ablation results for HTLA (first part) and results on AAPD and BGC datasets (second part) } \label{ablation}
\begin{tabularx}{1.0 \textwidth}{p{4.0cm} *{6}{>{\raggedright\arraybackslash}X}} % Adjust the number of columns as needed

%\begin{tabularx}{0.65 \textwidth}{p{2.4cm}{1.5cm}{1.5cm}{1.5cm}{1.5cm}{1.5cm}{1.5cm}}
\toprule
\multirow{2}{*}{Model} & \multicolumn{2}{c}{WOS} & \multicolumn{2}{c}{RCV1-V2} & \multicolumn{2}{c}{NYT} \\ \cmidrule(lr){2-7} 
                       & MiF1      & MaF1      & MiF1      & MaF1      & MiF1      & MaF1      \\ \midrule

w/o TLA(BERT-GPTrans) & 86.74 & 80.62 & 86.28 & 68.19 & 78.89 & 67.34 \\
w/o $embed_{name}$ & 86.37 & 80.51  & 86.71 & 68.10 & 78.87 & 67.21 \\
w/o $embed_{node}$  & 86.48 & 80.58  & 86.90 & 68.45 & 79.58 & 68.24 \\
w/o  $LE$ & 86.81 & 80.87  & 86.53 & 68.38 & 79.15 & 68.75\\
HTLA & \textbf{87.38} & \textbf{81.88} & \textbf{87.14} & \textbf{70.05} & \textbf{80.30} & \textbf{69.74} \\
\midrule
\multirow{2}{*}{Model} & \multicolumn{3}{c}{AAPD (2-level hierarchy)} & \multicolumn{3}{c}{BGC (4-level hierarchy)} \\ 
\cmidrule(lr){2-4} \cmidrule(lr){5-7}
                       & MiF1      & MaF1  &     & MiF1      & MaF1 &      \\ 
\midrule
BERT & 57.65 & 80.90 && 63.21  & 79.77  &\\
BERT-GPTrans  & 58.17 & 81.17 && 64.28 & 80.48 & \\
HTLA & \textbf{62.37}  & \textbf{81.95}  && \textbf{66.05} & \textbf{81.05} & \\
 \bottomrule
\end{tabularx}

\end{table}

\section{Conclusion}

Existing methods face challenges in effectively aligning text-label semantics within the hierarchy. To address this, we propose TLA, a loss function explicitly modeling the alignment between text and its associated labels. Building upon this, we introduce HTLA model, employing a two-encoder architecture to merge text-label embeddings for enhanced representations in HTC. Our experiments show HTLA outperforms existing methods on benchmark datasets. We further analyze its performance amid label imbalance, across hierarchy levels, and based on the number of label paths to demonstrate effectiveness. Additionally, we validate HTLA's components and generalization capabilities. In future work, we aim to extend our approach to non-textual domains like images, biological data, and other multi-modal datasets.

\appendix

\section{Details of statistical test} \label{stat}

We evaluated the effectiveness of our implemented models by analyzing Micro-F1 (MiF1) and Macro-F1 (MaF1) scores, reporting average results from 8 runs. Subsequently, we employed one-sided paired t-tests to assess the significance of performance variations among the models across the three datasets as detailed in Table \ref{ttest}. Except for the Micro-F1 score for the HTLA vs. HGCLR comparison in WOS, all p-values for comparisons are significantly below the threshold of 0.05, implying that the HTLA model demonstrates a statistically significant performance improvement.

\begin{table*}[t]
\renewcommand{\arraystretch}{1}
\centering
%\small
%\tiny

%\captionsetup{font=scriptsize}
\caption{p-value for one-sided t-test} \label{ttest}
\begin{tabularx}{\textwidth}{p{4cm} *{6}{>{\raggedright\arraybackslash}X}} % Adjust the number of columns as needed

\toprule

\multirow{2}{*}{Model} & \multicolumn{2}{c}{WOS} & \multicolumn{2}{c}{RCV1-V2} & \multicolumn{2}{c}{NYT} \\ \cmidrule{2-7}
                       & MiF1      & MaF1      & MiF1      & MaF1      & MiF1      & MaF1      \\ \midrule
            
HTLA vs HGCLR & 0.23 & \textbf{1.8e-4} & \textbf{3.7e-5}   & \textbf{1.8e-4} & \textbf{1.3e-6} & \textbf{3.4e-7} \\
HTLA vs BERT-GPTrans & \textbf{2.4e-2} & \textbf{4.2e-4} & \textbf{1.5e-6}   & \textbf{3.1e-5} & \textbf{5.1e-6} & \textbf{2.7e-7} \\
HTLA vs BERT         & \textbf{5.1e-3} & \textbf{1.7e-4} & \textbf{6.2e-8}  & \textbf{2.2e-6} & \textbf{4.5e-7} & \textbf{1.3e-8}  \\

 \bottomrule
\end{tabularx}
\end{table*}

\section{Performance analysis on additional datasets} \label{bgcapd}

We conducted experiments on two additional datasets, namely AAPD and BGC, to validate the generalization capabilities of the HTLA model. AAPD consists of abstracts of scientific papers from the arXiv.org\footnote{\url{https://arxiv.org/}} website, while BGC\footnote{\url{https://www.inf.uni-hamburg.de/en/inst/ab/lt/resources/data/blurb-genre-collection.html}} contains book blurbs from the Penguin Random House website. Both datasets consist of multipath labels. Table \ref{tabbgcapd} provides detailed statistics for the two datasets. 

\begin{table}[t]
    \centering
   % \scriptsize
    %\tiny
    %\captionsetup{font=scriptsize}
    \caption{Statistical details for the AAPD and BGC. $|Level|$ indicates the number of hierarchy levels, $|L|$ is the total label count, and Mean-$|L|$ denotes the mean number of labels per sample}
    \label{tabbgcapd}
    \renewcommand {\arraystretch}{1.2}
    \begin{tabular}{ccccccc} 
    \toprule
    Dataset & $|Level|$ & Train & Val & Test & $|L|$ & Mean-$|L|$ \\
    \midrule
    AAPD & 2 & 53840 & 1000 & 1000 & 61 & 4.09 \\
    BGC & 4 & 58715 & 14785 & 18394 & 146 & 3.01 \\
    \bottomrule
    \end{tabular}
\end{table}

\begin{credits}
\subsubsection{\ackname} This study was funded by the PMRF (Prime Minister's Research Fellow) program, run by the Ministry of Education, Government of India. We also acknowledge National Supercomputing Mission (NSM) for providing computing
resources of ‘PARAM Ganga’ at IIT Roorkee, which is implemented by C-DAC and supported
by the Ministry of Electronics and Information Technology (MeitY) and Department of
Science and Technology (DST), Government of India.

\subsubsection{\discintname}
The authors have no competing interests to declare that are
relevant to the content of this article.
\end{credits}
%% Note that this preceding line implies that you store your BibTeX references in a file called 'mybibliography.bib'. If you instead store your references in a file with a different name, for instance 'references.bib', the preceding line should read '\bibliography{references}'. Whatever you do, DO NOT put the file name extension .bib inside the \bibliography command; this will trip up LaTeX compilers. 

\begin{thebibliography}{10}
% Note that this number 8 reserves an amount of space (equal to the natural width of the given number) for the label of your references; if you have more than 9 references, you will want to change this number to 18. If you have more than 19 references, this number is best changed to 88. If you have more than 99 references, I salute you.



\providecommand{\url}[1]{\texttt{#1}}
\providecommand{\urlprefix}{URL }
\providecommand{\doi}[1]{https://doi.org/#1}

\bibitem{Aly2019}
Aly, R., Remus, S., Biemann, C.: Hierarchical multi-label classification of text with capsule networks. In: Proceedings of the 57th Annual Meeting of the Association for Computational Linguistics: Student Research Workshop. pp. 323--330. Association for Computational Linguistics, Florence, Italy (Jul 2019). \doi{10.18653/v1/P19-2045}, \url{https://aclanthology.org/P19-2045}

\bibitem{Chen2020}
Chen, B., Huang, X., Xiao, L., Cai, Z., Jing, L.: Hyperbolic interaction model for hierarchical multi-label classification. Proceedings of the AAAI Conference on Artificial Intelligence  \textbf{34}(05),  7496--7503 (Apr 2020). \doi{10.1609/aaai.v34i05.6247}, \url{https://ojs.aaai.org/index.php/AAAI/article/view/6247}

\bibitem{Chen2021}
Chen, H., Ma, Q., Lin, Z., Yan, J.: Hierarchy-aware label semantics matching network for hierarchical text classification. In: Proceedings of the 59th Annual Meeting of the Association for Computational Linguistics and the 11th International Joint Conference on Natural Language Processing (Volume 1: Long Papers). pp. 4370--4379. Association for Computational Linguistics, Online (Aug 2021). \doi{10.18653/v1/2021.acl-long.337}, \url{https://aclanthology.org/2021.acl-long.337}







\bibitem{ntxent2020}
Chen, T., Kornblith, S., Norouzi, M., Hinton, G.: A simple framework for contrastive learning of visual representations. In: Proceedings of the 37th International Conference on Machine Learning. ICML'20, JMLR.org (2020)

\bibitem{GPTrans2024}
Chen, Z., Tan, H., Wang, T., Shen, T., Lu, T., Peng, Q., Cheng, C., Qi, Y.: Graph propagation transformer for graph representation learning. In: Elkind, E. (ed.) Proceedings of the Thirty-Second International Joint Conference on Artificial Intelligence, {IJCAI-23}. pp. 3559--3567. International Joint Conferences on Artificial Intelligence Organization (8 2023). \doi{10.24963/ijcai.2023/396}, \url{https://doi.org/10.24963/ijcai.2023/396}, main Track

\bibitem{CUNHA2021}
Cunha, W., Mangaravite, V., Gomes, C., Canuto, S., Resende, E., Nascimento, C., Viegas, F., França, C., Martins, W.S., Almeida, J.M., Rosa, T., Rocha, L., Gonçalves, M.A.: On the cost-effectiveness of neural and non-neural approaches and representations for text classification: A comprehensive comparative study. Information Processing \& Management  \textbf{58}(3),  102481 (2021). \doi{https://doi.org/10.1016/j.ipm.2020.102481}, \url{https://www.sciencedirect.com/science/article/pii/S0306457320309705}

\bibitem{Deng2021}
Deng, Z., Peng, H., He, D., Li, J., Yu, P.: {HTCI}nfo{M}ax: A global model for hierarchical text classification via information maximization. In: Proceedings of the 2021 Conference of the North American Chapter of the Association for Computational Linguistics: Human Language Technologies. pp. 3259--3265. Association for Computational Linguistics, Online (Jun 2021). \doi{10.18653/v1/2021.naacl-main.260}, \url{https://aclanthology.org/2021.naacl-main.260}

\bibitem{Devlin2019}
Devlin, J., Chang, M.W., Lee, K., Toutanova, K.: {BERT}: Pre-training of deep bidirectional transformers for language understanding. In: Proceedings of the 2019 Conference of the North {A}merican Chapter of the Association for Computational Linguistics: Human Language Technologies, Volume 1 (Long and Short Papers). pp. 4171--4186. Association for Computational Linguistics, Minneapolis, Minnesota (Jun 2019). \doi{10.18653/v1/N19-1423}, \url{https://aclanthology.org/N19-1423}

\bibitem{dumais2000}
Dumais, S., Chen, H.: Hierarchical classification of web content. In: Proceedings of the 23rd Annual International ACM SIGIR Conference on Research and Development in Information Retrieval. p. 256–263. SIGIR '00, Association for Computing Machinery, New York, NY, USA (2000). \doi{10.1145/345508.345593}, \url{https://doi.org/10.1145/345508.345593}

\bibitem{gopal2013}
Gopal, S., Yang, Y.: Recursive regularization for large-scale classification with hierarchical and graphical dependencies. In: Proceedings of the 19th ACM SIGKDD International Conference on Knowledge Discovery and Data Mining. p. 257–265. KDD '13, Association for Computing Machinery, New York, NY, USA (2013). \doi{10.1145/2487575.2487644}, \url{https://doi.org/10.1145/2487575.2487644}

\bibitem{huang2019}
Huang, W., Chen, E., Liu, Q., Chen, Y., Huang, Z., Liu, Y., Zhao, Z., Zhang, D., Wang, S.: Hierarchical multi-label text classification: An attention-based recurrent network approach. In: Proceedings of the 28th ACM International Conference on Information and Knowledge Management. p. 1051–1060. CIKM '19, Association for Computing Machinery, New York, NY, USA (2019). \doi{10.1145/3357384.3357885}, \url{https://doi.org/10.1145/3357384.3357885}

\bibitem{Hull1993}
Hull, D.: Using statistical testing in the evaluation of retrieval experiments. In: Proceedings of the 16th Annual International ACM SIGIR Conference on Research and Development in Information Retrieval. p. 329–338. SIGIR '93, Association for Computing Machinery, New York, NY, USA (1993). \doi{10.1145/160688.160758}, \url{https://doi.org/10.1145/160688.160758}

\bibitem{Kowsari2017}
Kowsari, K., Brown, D.E., Heidarysafa, M., Jafari~Meimandi, K., Gerber, M.S., Barnes, L.E.: Hdltex: Hierarchical deep learning for text classification. In: 2017 16th IEEE International Conference on Machine Learning and Applications (ICMLA). pp. 364--371 (2017). \doi{10.1109/ICMLA.2017.0-134}

\bibitem{Lewis2004}
Lewis, D.D., Yang, Y., Rose, T.G., Li, F.: Rcv1: A new benchmark collection for text categorization research. J. Mach. Learn. Res.  \textbf{5},  361–397 (dec 2004)

\bibitem{lse-hiagm2024}
Liu, H., Huang, X., Liu, X.: Improve label embedding quality through global sensitive gat for hierarchical text classification. Expert Systems with Applications  \textbf{238},  122267 (2024). \doi{https://doi.org/10.1016/j.eswa.2023.122267}, \url{https://www.sciencedirect.com/science/article/pii/S0957417423027690}

\bibitem{Mao2019}
Mao, Y., Tian, J., Han, J., Ren, X.: Hierarchical text classification with reinforced label assignment. In: Proceedings of the 2019 Conference on Empirical Methods in Natural Language Processing and the 9th International Joint Conference on Natural Language Processing (EMNLP-IJCNLP). pp. 445--455. Association for Computational Linguistics, Hong Kong, China (Nov 2019). \doi{10.18653/v1/D19-1042}, \url{https://aclanthology.org/D19-1042}

\bibitem{umpmg2023}
Ning, B., Zhao, D., Zhang, X., Wang, C., Song, S.: Ump-mg: A uni-directed message-passing multi-label generation model for hierarchical text classification. Data Science and Engineering  \textbf{8},  1--12 (04 2023). \doi{10.1007/s41019-023-00210-1}

\bibitem{Peng2016}
Peng, S., You, R., Wang, H., Zhai, C., Mamitsuka, H., Zhu, S.: {DeepMeSH: deep semantic representation for improving large-scale MeSH indexing}. Bioinformatics  \textbf{32}(12),  i70--i79 (06 2016). \doi{10.1093/bioinformatics/btw294}, \url{https://doi.org/10.1093/bioinformatics/btw294}

\bibitem{Sandhaus2008}
Sandhaus, E.: {The New York Times Annotated Corpus - Linguistic Data Consortium}. The New York Times  (2008), \url{https://catalog.ldc.upenn.edu/LDC2008T19}

\bibitem{shen2021}
Shen, J., Qiu, W., Meng, Y., Shang, J., Ren, X., Han, J.: {T}axo{C}lass: Hierarchical multi-label text classification using only class names. In: Toutanova, K., Rumshisky, A., Zettlemoyer, L., Hakkani-Tur, D., Beltagy, I., Bethard, S., Cotterell, R., Chakraborty, T., Zhou, Y. (eds.) Proceedings of the 2021 Conference of the North American Chapter of the Association for Computational Linguistics: Human Language Technologies. pp. 4239--4249. Association for Computational Linguistics, Online (Jun 2021). \doi{10.18653/v1/2021.naacl-main.335}, \url{https://aclanthology.org/2021.naacl-main.335}

\bibitem{Wang2022}
Wang, Z., Wang, P., Huang, L., Sun, X., Wang, H.: Incorporating hierarchy into text encoder: a contrastive learning approach for hierarchical text classification. In: Proceedings of the 60th Annual Meeting of the Association for Computational Linguistics (Volume 1: Long Papers). pp. 7109--7119. Association for Computational Linguistics, Dublin, Ireland (May 2022). \doi{10.18653/v1/2022.acl-long.491}, \url{https://aclanthology.org/2022.acl-long.491}

\bibitem{hugggingfacetransformers}
Wolf, T., Debut, L., Sanh, V., Chaumond, J., Delangue, C., Moi, A., Cistac, P., Rault, T., Louf, R., Funtowicz, M., Davison, J., Shleifer, S., von Platen, P., Ma, C., Jernite, Y., Plu, J., Xu, C., Le~Scao, T., Gugger, S., Drame, M., Lhoest, Q., Rush, A.: Transformers: State-of-the-art natural language processing. In: Proceedings of the 2020 Conference on Empirical Methods in Natural Language Processing: System Demonstrations. pp. 38--45. Association for Computational Linguistics, Online (Oct 2020). \doi{10.18653/v1/2020.emnlp-demos.6}, \url{https://aclanthology.org/2020.emnlp-demos.6}

\bibitem{wu2019}
Wu, J., Xiong, W., Wang, W.Y.: Learning to learn and predict: A meta-learning approach for multi-label classification. In: Proceedings of the 2019 Conference on Empirical Methods in Natural Language Processing and the 9th International Joint Conference on Natural Language Processing (EMNLP-IJCNLP). pp. 4354--4364. Association for Computational Linguistics, Hong Kong, China (Nov 2019). \doi{10.18653/v1/D19-1444}, \url{https://aclanthology.org/D19-1444}

\bibitem{aapd2018}
Yang, P., Sun, X., Li, W., Ma, S., Wu, W., Wang, H.: {SGM}: Sequence generation model for multi-label classification. In: Bender, E.M., Derczynski, L., Isabelle, P. (eds.) Proceedings of the 27th International Conference on Computational Linguistics. pp. 3915--3926. Association for Computational Linguistics, Santa Fe, New Mexico, USA (Aug 2018), \url{https://aclanthology.org/C18-1330}

\bibitem{capsule2023}
Zhao, F., Wu, Z., He, L., Dai, X.Y.: Label-correction capsule network for hierarchical text classification. IEEE/ACM Transactions on Audio, Speech, and Language Processing  \textbf{31},  2158--2168 (2023). \doi{10.1109/TASLP.2023.3282099}

\bibitem{Zhao2021}
Zhao, R., Wei, X., Ding, C., Chen, Y.: Hierarchical multi-label text classification: Self-adaption semantic awareness network integrating text topic and label level information. In: Qiu, H., Zhang, C., Fei, Z., Qiu, M., Kung, S.Y. (eds.) Knowledge Science, Engineering and Management. pp. 406--418. Springer International Publishing, Cham (2021)

\bibitem{Zhou2020}
Zhou, J., Ma, C., Long, D., Xu, G., Ding, N., Zhang, H., Xie, P., Liu, G.: Hierarchy-aware global model for hierarchical text classification. In: Proceedings of the 58th Annual Meeting of the Association for Computational Linguistics. pp. 1106--1117. Association for Computational Linguistics, Online (Jul 2020). \doi{10.18653/v1/2020.acl-main.104}, \url{https://aclanthology.org/2020.acl-main.104}

\end{thebibliography}
%
% If you do not want to use BibTeX, you can also type up the bibliography exactly as you see fit, using the following structure:

\end{document}